# Towards Resilient Artificial Intelligence: Survey and Research Issues

Oliver Eigner, Sebastian Eresheim, Peter Kieseberg, Lukas Daniel Klausner,
Martin Pirker, Torsten Priebe, Simon Tjoa
Institute of IT Security Research
St. Pölten University of Applied Sciences
{oliver.eigner, sebastian.eresheim, peter.kieseberg, lukas.daniel.klausner,
martin.pirker, torsten.priebe, simon.tjoa}@fhstp.ac.at

Fiammetta Marulli
Department of Mathematics and Physics
University of Campania "Luigi Vanvitelli"
fiammetta.marulli@unicampania.it

Francesco Mercaldo
Department of Medicine and
Health Sciences "Vincenzo Tiberio"
University of Molise
francesco.mercaldo@unimol.it

*Abstract*—Artificial intelligence (AI) systems are becoming critical components of today's IT landscapes. Their resilience against attacks and other environmental influences needs to be ensured just like for other IT assets. Considering the particular nature of AI, and machine learning (ML) in particular, this paper provides an overview of the emerging field of resilient AI and presents research issues the authors identify as potential future work.

## I. Introduction

AI systems are becoming indispensable for our daily lives. Organisations should ensure the resilience of such systems, just as they would for any other critical asset. However, the "black box" approach typical for AI may make assessing and ensuring resilience different when compared to traditional IT systems. This position paper aims to provide an overview of the emerging field of resilient AI research, with the goal to identify issues the authors intend to work on in the future. Methodologically, we pursue this by an in-depth literature survey and categorisation by both application area and technique.

To this end, we first need to define what "resilient AI" means. We would like to point out that technological systems can never be regarded as isolated, abstracted entities, but must be seen in their societal context as socio-technical systems; the understanding must be that algorithms, and AI in particular, are not just technical artefacts (in the sense of "physical objects designed by humans that have both a function and a use plan", as defined by Vermaas et al. [1]), but complex systems, shaped by collective and distributive agency [2]. In our case, this means that the concept of "resilient AI" has links to questions of e. g. acceptance and trust, and consequently connections to the classical FAccT triad (fairness, accountability and transparency): No AI system can be considered resilient if its use is fraught with fundamental mistrust, responsibility vacuums, (nearly always justified) accusations of injustice or critique of its black-box-ness, to mention only the most prominent aspects. Similarly, the EU's JRC differentiates between four dimensions of resilience with regard to critical infrastructure: societal, economic, organisational and technological [3].

However, having pointed out these more encompassing and broader conceptions of resilience, in this short position paper we will focus on the more technical aspects. In particular, we point to the various definitions of "resilience" given by the National Institute of Standards and Technology (NIST),[1] which we synthesise as follows: Resilience (in its more technical sense) is the ability of an information system to reduce the magnitude, impact and/or duration of disruptive events or, more generally, any known or unknown changes in the operating environment (including deliberate attacks, accidents and naturally occurring threats or incidents) by a) anticipating and preparing for such events (through e. g. risk management, contingency and continuity planning), b) being able to withstand and adapt to attacks, adverse conditions or other stress and potential disruptions, and continuing (or rapidly recovering the ability) to operate (even if in a degraded or debilitated state) while maintaining essential and required operational capabilities, and c) recovering full operational capabilities after such a disruption in a time frame consistent with mission needs.

The remainder of this paper is structured as follows: Sections II and III present our literature survey of related work based on two dimensions. Section II focuses on several application areas in which resilience of AI systems is particularly important. Section III then addresses the topic from the perspective of techniques which may pose particular threats to the resilience of AI systems. These two dimensions are also shown below in figure 1. Based on this literature survey, in section IV we identify research issues which we would like to pursue moving forward. Finally, section V concludes the paper with a brief summary.

---

[1] https://csrc.nist.gov/glossary/term/resilience (accessed 2021-05-03)

## II. Application Areas

**Military.** Many countries adopt AI into their military doctrines, focusing on different application areas ranging from drone swarms to predictive maintenance and information analysis. In using AI (as in all military applications), money is key in defining the capabilities [4], thus requiring smaller military forces to be highly selective in their investments. This is especially important when considering the problem of missing explainability in many advanced AI methods [5], i. e. the fact that decision-making processes inside these applications are "black boxes" (for which testing for backdoors is virtually impossible, thus requiring the military to ultimately trust the vendors [6]). Furthermore, data used for training AI-based systems can quickly become obsolete, thus requiring advanced strategies for keeping decision-making up to date [7]. Additionally, like any new system introduced into the war theatre, AI-based methods introduce a new attack surface and thus allow for novel attacks. In particular, the issue of adversarial machine learning (AML) receives specific attention in this domain [8].

**Policing and Justice.** The broad area of law enforcement, policing and justice is a particularly sensitive field, especially concerning the use of algorithms, automation and AI. Numerous different kinds of automation have been or are being employed in this sector [9], such as recidivism prediction [10] or predictive policing algorithms [11], to give just two prominent and controversial examples. The increasing use of digitisation, automation and datafication already raises a broad array of difficult questions even without considering the specific question of resilience, adversarial attacks and disruptions to critical systems; however, it is clear that resilience and withstanding disruptive and hostile events is of particular concern in a delicate area so closely concerned with civil and human rights as well as the rule of law. This is also reflected in recent calls for research proposals by the European Commission.[2]

**Health.** In the health area, computer-aided diagnosis (CAD) systems trained using large amounts of patient data (mostly physiological signals and images based on integration of advanced signal processing and machine learning techniques in an automated fashion) can assist neurologists, neurosurgeons, radiologists and other medical providers to make better clinical decisions. Clinical medicine has emerged as a promising application area for AI and AI has already achieved human-level performance in clinical pathology [12], dermatology [13], radiology [14], neurology [15] as well as ophthalmology [16]. Research on secure and robust machine learning in this area has been growing at an accelerating pace in the past decade [17].

**Industry.** In the context of Industry 4.0 and robotics, the notion of "digital twins" has emerged to allow simulation and optimisation of production and other industrial processes. Even without the presence of deliberate attacks, the operating environment (e. g. physical parameters) may change, leading to a particular need of resilience and robustness [18]. Adversarial attacks have also been carried out against trajectory planning systems of robots [19]. Likewise, SCADA is a control system architecture for high-level process supervisory management in industrial environments. Several factors have contributed to the escalation of risks specific to SCADA infrastructure systems, including the adoption of standardised technology with known vulnerabilities, interconnectivity with other networks and use of insecure remote connections. The main concern is that these systems are usually evaluated only for a limited set of well-known attacks, meaning they are particularly vulnerable against zero-day threats. Recently, poisoning and adversarial attacks have also been successfully performed against SCADA systems [20].

**Finance.** The financial sector has been heavily influenced by the digital transformation in the recent past. AI applications represent a great opportunity to make financial organisations more competitive and profitable; however, this development is also accompanied by manifold risks. Firstly, organisations in the financial sector are by their nature a particularly attractive target for cybercriminals. Secondly, they must meet (with good reason) particularly high compliance requirements. The importance of cyber resilience is emphasised all over the world by key institutions such as the ECB [21] or the BIS [22]. Carminati et al. [23] applied adversarial ML attacks on fraud detection algorithms used by organisations in the banking sector in an experimental setting. Opening an account online, for example, could pose a problem in the future if deep fakes are used to impersonate another person – in particular if an account created this way is then be used for money laundering. Another possible problem area is that AI applications might accidentally form a cartel by adapting their behaviour to conform to other AI applications in the same sector.

**Security.** There have been several successful applications of DL methods in the cyber security field [24], particularly in malware detection [25]. Typically, these methods involve classifiers focused on identifying malicious behaviour or vulnerabilities. Criticism has recently been raised against these systems as they do not provide any kind of explanation and interpretation for the generated predictions. For this reason, there is a need not only to design accurate vulnerability detection systems, but also to provide an accompanying rationale [26]; in fact, if users are more aware of the reasons for a prediction, they are also more likely to follow recommended solutions. Considering that these methods are usually evaluated on old threats, however, the resilience of such systems (for instance against zero-day attacks) is rarely tested, which also contributes to lack of user confidence in them. For all of these reasons, the introduction of resilience into cyber security is necessary to increase trust [27, 28].

---

[2] https://ec.europa.eu/info/funding-tenders/opportunities/portal/screen/opportunities/topic-details/su-ai02-2020 (accessed 2021-05-03)

## III. AI Techniques

**Data Preparation and Management.** AI systems consume data, parse and make sense of it, and then act thereon. Consequently, the pathway of data ingestion also needs consideration in terms of resilience. If data is prepared and ingested in the expected format, all works well. However, if the incoming data is malformed, an AI implementation may end up in an unexpected state it was never intended to be in. It becomes a "weird machine" – to use a term originating from exploits in security research [29], where data fed to a system is deliberately, maliciously manipulated to trigger weird, unexpected behaviour(s). Besides the obvious threat of deliberate attacks, even data ingestion in common formats is open to accidental failure (e. g. for XML or JSON [30]) as parser implementations do not follow identical specifications. Extrapolating from this, media formats (audio or video) are even more complex – and they do sometimes fail even in popular platforms [31]. To counteract such deficiencies, there are efforts to reimplement data ingestion parsers more robustly [32]. But to create a resilient, hardened AI system, each data preparation, ingestion and internal processing component needs to be evaluated according to known robustness issues in implementation engineering. Paz et al. [33] present an approach for resilient data management in streaming data processing for the internet of things by adding a fault tolerance strategy addressing connectivity issues of mobile devices to the NebulaStream platform.

**Deep Learning.** Deep learning (DL) is a kind of ML characterised by its organisation of knowledge in hierarchies of concepts many layers deep [34]. Deep Neural Networks (DNN) represent its most popular family of algorithms. DL approaches have been successfully adopted for several fields and applications; unfortunately, DL models have consequently become targets of attacks, which differentiated by a) the attacker's goals (espionage, sabotage, fraud) and b) the stages of ML pipeline attacked (training or production); such attacks are called *attacks on algorithms* and *attacks on models*, respectively. Common DL attacks are *evasion, poisoning, trojaning, backdooring, reprogramming and inference* attacks. Evasion is the most common type of attack on DL models: It is performed during production and exploits one major intrinsic weakness, the vulnerability to *adversarial examples* (AE) [35], which are specific data samples able to influence a learning model and corrupt its inference capabilities. Indeed, the adversarial machine learning (AML) research strand has flourished [36], leading to Generative Adversarial Networks (GANs) [37], a kind of DNN aiming to artificially generate ad-hoc adversarial data for attacking DL-based systems. One prominent example among DL applications is natural language processing (NLP), which is particularly vulnerable to data poisoning attacks performed on linguistic compressed distributional models (best-known as *word embeddings* (WEs) [38]). Poisoning attacks on WEs [39] aim to modify a model's perceived meaning of new and existing words by changing their locations in the embedding space, thus exerting control on the DNN models.

**Reinforcement Learning.** Huang et al. [40] provided the first examples that adversarial attacks are also possible on reinforcement learning (RL) agents. They showed that applying the same perturbations as in supervised learning on images leads to a significant drop in policy performance across different deep reinforcement learning (DRL) algorithms. However, they assume the adversary can manipulate the full input data at every time step and knows about the internal parameters of the victim's policy. Hussenot et al. [41] extend these attacks to change the victim's behaviour in a specific direction rather than just a more general distortion. They again assume a white-box approach and control of the full input data. Behzadan and Munir [42] accomplish a similar result without relying on the victim's policy parameters. However, the other two assumptions are still required. Under similar assumptions Pattanaik et al. [43] influence an agent to always take the worst possible action. Interestingly, training an agent using such attacks leads to better robustness in environments with different physical properties. Pinto et al. [44] take this approach a step further by modeling uncertainties of physical properties in simulation environments via an adversarial agent that applies disturbance forces in order to make the agent policy more robust. In contrast to the previously mentioned work, Qu et al. [45] only assume a black-box approach (only the input state and output distribution of an agent's policy are known), control over fractional input data and only sporadic disturbance. Gleave et al. [46] point out that a performance reduction is even possible when the adversary is only a part of the victim's environment and therefore part of its observations. They also argue that higher-dimensional states are more susceptible to such attacks. However, the adversary is primarily trained to win in zero-sum games against pre-trained victims. The disturbance is caused by the adversary exploiting the victim's incomplete understanding of its environment. A possible countermeasure against such vulnerabilities could be league training, introduced by Vinyals et al. [47]. Huai et al. [48] demonstrate that not only policies can be the target of adversaries, but also the interpretation methods which try to explain the reasoning behind chosen actions. Additionally, they also describe a model poisoning attack targeting these interpretations. Recently, Yang et al. [49] proposed a robust learning scheme based on causal inference. They showed that an agent exposed to disturbances and accompanying labels is more robust against adversarial disturbances afterwards.

**Federated Learning.** Federated learning (FL) [50] is a novel distributed approach for performing heavy computations without moving data from its original locations. In a FL-based task, a common learning model is distributed to all the participants (clients), each of whom locally trains a shared model on its own data fragment; then, they send this model back to a central node that serves as the local models' aggregator. This aggregator then merges local models into a global model (using a federated average or consensus, for example), which is then sent back to each client to further proceed in the task. FL has the potential of mitigating, if only partially, data protection problems: Local data are not shared, and only locally trained models and the global model are shared among the federation. Furthermore, confidentiality of the training data is protected by limiting the shared information between server and clients only to model parameters. However, this setting is vulnerable to several adversarial attacks [51], performed by injecting malicious training updates with the goal of influencing the main federated task. Notable examples of adversarial algorithms are targeted attacks (e. g. backdoor attacks, model and data poisoning attacks [52]) and untargeted attacks (e. g. Byzantine failures [53]).

**Model Evaluation and Operations.** Traditionally, ML models are evaluated using metrics such as accuracy or AUC. Such metrics, however, only measure the performance of a model on the utilised test data and do not consider its performance on a wider range of data (including "under attack") and over time. Therefore, resilience measures have been proposed, e. g. by Katzir and Elovici [54]. In particular, ML models degrade over time. The reason for this is the fact that the world is not static; hence the data to which the models are applied also change over time. This effect is called "concept drift". Drift detection methods have been discussed for several years [55], but appropriate monitoring of AI systems has only recently received attention with emerging trends such as MLOps.

## IV. RESEARCH ISSUES AND FUTURE WORK

Based on the above literature survey, we have identified the following research issues as shown in Figure 1. This is, however, by no means an exhaustive list of open research topics in the area of resilient AI, but rather points towards intended future work of the authors of this paper.

**Legal Aspects in Data Cleansing.** One major issue that has thus far received almost no attention is the impact of data cleansing on AI-based systems, especially concerning more complex methods that currently have an explainability problem; here, the actual impact of data cleansing cannot be estimated in most cases. On the other hand, data cleansing is mandatory in almost all real-world applications, as faulty or incomplete data is the norm even in strictly deterministic systems. Besides the more technical problems associated with data cleansing, there is also the question of legal issues from the cleansing process – data cleansing typically requires changes to the data, either through removal of the data records in question or by changing values to defaults. Due to the explainability problem, even proper annotation of the changes made does not fully explain the impact on AI-based decisions; thus the question of responsibility for errors and negative effects introduced into the end results through these changes needs to be discussed.

**Adversarial Policies in Reinforcement Learning.** Most current adversarial attacks in RL target the DQN algorithm in combination with images as the dominant environment state factor. Also, often the same supervised DL attacks are applied to RL at multiple time steps, which by design ignore the agent's sequential decision-making process. Huang et al. [40] already pointed out that from the three algorithms they experimented with, TRPO and A3C seem to be more resistant than DQN by default against supervised DL adversarial attacks. Since the literature following the paper of Huang et al. [40] mostly focuses on DQN, the question arises whether other RL algorithms are susceptible to more advanced adversarial attacks which *do* build on the agent's sequential decision-making process. Furthermore, most prior work requires quite strong assumptions, making them theoretically possible, yet practically challenging. In our opinion, Gleave et al. [46] present the most realistic scenario: In a multi-agent environment, the adversary is itself an agent and thus part of the target agent's environment. The adversary is therefore only able to influence the target agent's input partially (and not arbitrarily), will likely not influence the agent on every time step and is uninformed about the target agent's inner decision-making process. In this setting, targeted attacks where the adversary successfully causes the target agent to act in a very specific, predefined way remain to be seen. On the countermeasure side, Yang et al. [49] introduce causal inference for resilient RL agents in a very specific setting. We think that a combination of these two research topics holds great potential in order to build RL agents robust against a wide range of adversarial attacks.

**Resilience Testing and Monitoring.** Penetration testing is a key element to assess the security and resilience of ICT services and products. As myriads of threats [56] in the area of AI exist which are capable of negatively affecting the correct operations of AI applications, AI pentesting seems certain to gain importance in the new future. First frameworks and approaches (e. g. [6, 57]) have already been proposed or are currently under development. However, we still see a large gap in the area of AI pentesting tools and specific approaches using an adversarial perspective to test AI applications. In addition, measuring and monitoring resilience against even non-adversarial changes of the environment opens up yet another area of research. We see the need for a holistic view on AI audit and certification [58].

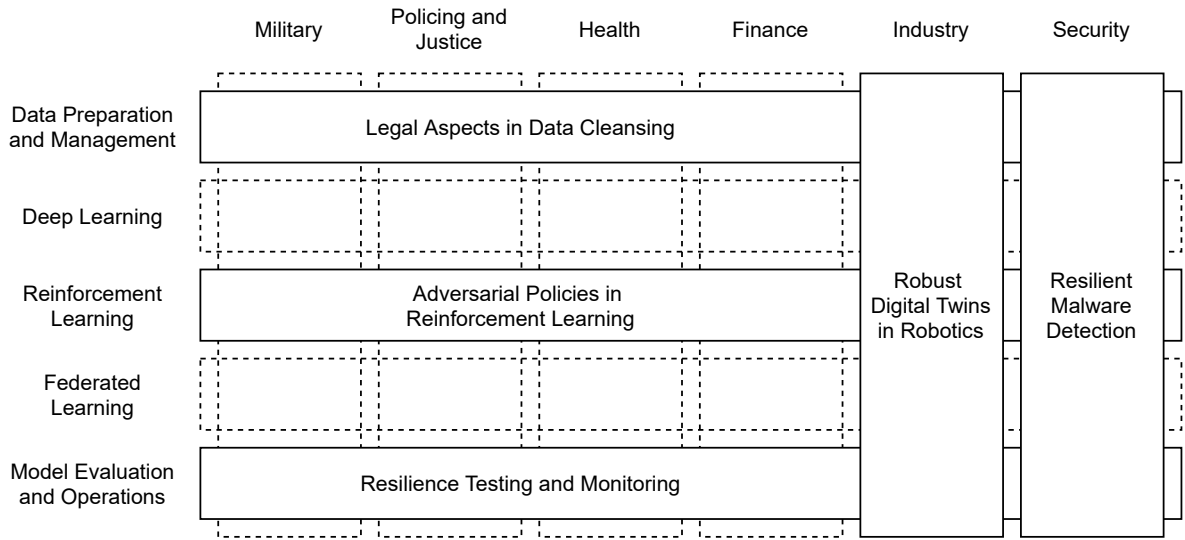

Figure 1: Survey dimensions and identified research issues

**Robust Digital Twins in Robotics.** As pointed out in section II, a major challenge in robotics is to make "digital twins" robust against changing parameters of the environment. As Pattanaik et al. [43] have shown, being robust against adversarial attacks does not only mean being robust against an adversary, but also being robust against (unexpected) variations of parameters in the environment. These approaches can be seen as specific examples of *domain randomisation* [59, 60]. We see this as a promising approach to bridge the gap between simulation and real-world applications.

**Resilient Malware Detection.** The traditional approach in anti-malware software is *signature-based*: Essentially, code snippet signatures gathered from real-world malicious samples are used by means of pattern-matching heuristics to classify code as either benign or malicious. Unfortunately, the application of code mutation [61] or obfuscation techniques [62] allows malicious code to evade detection. Therefore, malware detectors must also include obfuscated samples in their testing to achieve resilience against these techniques. Given the rising challenges in direct code analysis of malware, a competing approach is a *system-centric* one: The characteristic interactions of (benign) programs with operating system resources [63] or system-wide process fingerprinting for anomalous behaviours [64] is used as the key differentiator to detect malware. Novel attacks can be expected, by definition, to also exhibit novel, anomalous behaviour. The challenge thus lies in whether trained, system-wide observers of processes are sufficiently resilient to new variations of malicious actions.

## V. Conclusion

The goal of this paper was to provide an overview of current resilient AI research and to identify open research issues. We addressed this by structuring the field in two dimensions: application areas in which resilience of AI systems is particularly important as well as AI techniques which may pose particular threats to the resilience of AI systems. We point out once more that the identified issues focus on intended future work of the authors and are not meant to be an exhaustive list of open research topics in the field.